\documentclass[review,12pt, 3p]{elsarticle}

\usepackage{wrapfig}
\usepackage{graphicx}
\usepackage{setspace} 

\usepackage{amssymb}
\usepackage{amsmath}
\usepackage{multirow}
\usepackage{hyperref}
\biboptions{sort&compress}
\usepackage[caption=false,font=normalsize,labelfont=sf,textfont=sf]{subfig}

\usepackage[figuresright]{rotating}

\begin{document}

\begin{frontmatter}




\title{HTNet for micro-expression recognition}


 \author[rvt]{Zhifeng Wang\corref{cor1}}
 \ead{zhifeng.wang@anu.edu.au}
 \author[rvt]{Kaihao Zhang }
 \ead{super.khzhang@gmail.com}
 \author[focal]{ Wenhan Luo}
 \ead{whluo.china@gmail.com}
 \author[rvt]{Ramesh Sankaranarayana\corref{cor2}}
 \ead{ramesh.sankaranarayana@anu.edu.au}

\cortext[cor1]{Principal  corresponding author}
\cortext[cor2]{Corresponding author} 
 \address[rvt]{College of Engineering and Computer Science, Australian National University, Canberra, ACT, Australia}
 \address[focal]{Sun Yat-sen University, Guangzhou, China}

\begin{abstract}
Facial expression is related to facial muscle contractions and different muscle movements correspond to different emotional states.  For micro-expression recognition, the muscle movements are usually subtle, which has a negative impact on the performance of current facial emotion recognition algorithms.  Most existing methods use self-attention mechanisms  to capture relationships between tokens in a sequence, but they do not take into account the inherent spatial relationships between facial landmarks. This can result in sub-optimal performance on  micro-expression recognition tasks.Therefore, learning to recognize facial muscle movements is a key challenge in the area of micro-expression recognition.  In this paper, we propose a Hierarchical Transformer Network (HTNet) to identify critical areas of facial muscle movement.  HTNet includes two major components: a transformer layer that leverages the local temporal features and an aggregation layer that extracts local and global semantical facial features.  Specifically, HTNet divides the face into four different facial areas: left lip area, left eye area, right eye area and right lip area.  The transformer layer is used to focus on representing local minor muscle movement with local self-attention in each area.  The aggregation layer is used to learn the interactions between eye areas and lip areas. The experiments on four publicly available micro-expression datasets show that the proposed approach outperforms previous methods by a large margin. The codes and models are available at: \url{https://github.com/wangzhifengharrison/HTNet}
\end{abstract}


\end{frontmatter}


\section{Introduction}
Micro-expression refers to subtle muscle movements that last only for approximately 0.04 seconds. In recent years, extensive research has been conducted on utilizing computer vision-based methods to analyze micro-expressions \cite{ref-47}. However, the accuracy of recognizing micro-expressions still requires improvement. Even though the micro-expression datasets are collected in well-controlled laboratory environments, the current results are still unsatisfactory \cite{ref-9}. Utilizing computer vision for this task remains challenging due to the presence of subtle muscle movements that often accompany micro-expressions, making them difficult for both humans and computers to detect. On the other hand, the recognition of normal macro expressions has achieved high accuracy rates (over 95\%) \cite{ref-4}. This stark difference in performance can be attributed to the fact that micro-expressions are incredibly hard to detect due to their fleeting nature and subtle characteristics. As a result, researchers are still working to improve the precision and reliability of micro-expression recognition using computer vision techniques.\\
In the domain of micro-expression recognition, several researchers have proposed the adoption of the Local Binary Pattern (LBP) method to extract facial features. The LBP technique has demonstrated commendable discrimination ability through its texture-based feature extraction approach, while maintaining a low computational complexity \cite{ref-8}. Moreover, other scholars have explored the use of optical flow features as inputs for estimating muscle motion \cite{ref-8}. Optical flow is derived from the differences in brightness between consecutive frames, enabling the estimation of subtle facial movements. Several optical flow-based methodologies have been investigated, including Bi-WOOF \cite{ref-24}, MDMO \cite{ref-26}, FHOFO \cite{ref-25}, Optical Strain Weight, and Optical Strain Feature \cite{ref-27}.Furthermore, notable deep learning models such as VGG16 \cite{ref-14}, GoogleNet \cite{ref-13}, AlexNet \cite{ref-11}, and OFF-Apex \cite{ref-9} have been utilized to process the TV-L1 optical flow. This optical flow is extracted from selected apex and onset frames of the image sequences. The peak frame, capturing the most salient information about the micro-expression, proves to be instrumental in accurately recognizing the expression.\\
The recognition of micro-expressions involves two primary stages: feature classification of facial features and the extraction of relevant facial information from images. Traditional classification methods often rely on artificially generated features, while deep learning techniques automatically detect features from facial images. To address the challenge of understanding subtle facial movements and the limited availability of training data, Liu \textit{et al.} \cite{ref-8} propose using domain adaptation methods to transfer macro-expression knowledge for micro-expression recognition. This involves magnifying micro-expressions and augmenting the training dataset with more generated images, known as expression magnification and reduction (EMR). The system's performance using this approach proves to be competitive in the Micro-Expression Grand Challenge. Another approach suggested by Liong \textit{et al.} is the use of STSTNet, a three shallow stream CNN, to recognize facial expressions. The optical strain, vertical optical flow, and horizontal optical flow fields are fed into these three shallow stream networks \cite{ref-5}. For recognizing micro-expressions, Xia \textit{et al.} propose adopting a recurrent convolutional network, which has a shallower design and combines the strengths of both RNNs and CNNs \cite{ref-6}. Detecting micro-expressions is challenging due to the small and subtle nature of the facial movements. This makes it difficult to identify the specific facial muscles involved in the expression and monitor the pixel's motion over time. To address these issues, Zhou \textit{et al.} \cite{ref-15} suggest using a feature refinement network, specific expression feature learning network, and fusion techniques for expression recognition. By employing self-attention on global image features, their method can extract distinguishing characteristics for micro-expression recognition. However, global self-attention between image features will lack of fine-grained features. To deal with this issue,  our objective is to maintain self-attention in the local blocks at each hierarchy level to capture fine-grained features. We use aggregation blocks to incorporate both local fine-grained and global coarse-grained interactions at different image scales. With this new mechanism, each pixel in same-level blocks is treated at a fine granularity, while the pixels in upper-level blocks are at a coarse granularity. This allows our model to effectively capture both short- and long-range visual dependencies.\\
  \begin{figure*}[t]
 \centering
  \includegraphics[width=0.6\linewidth]{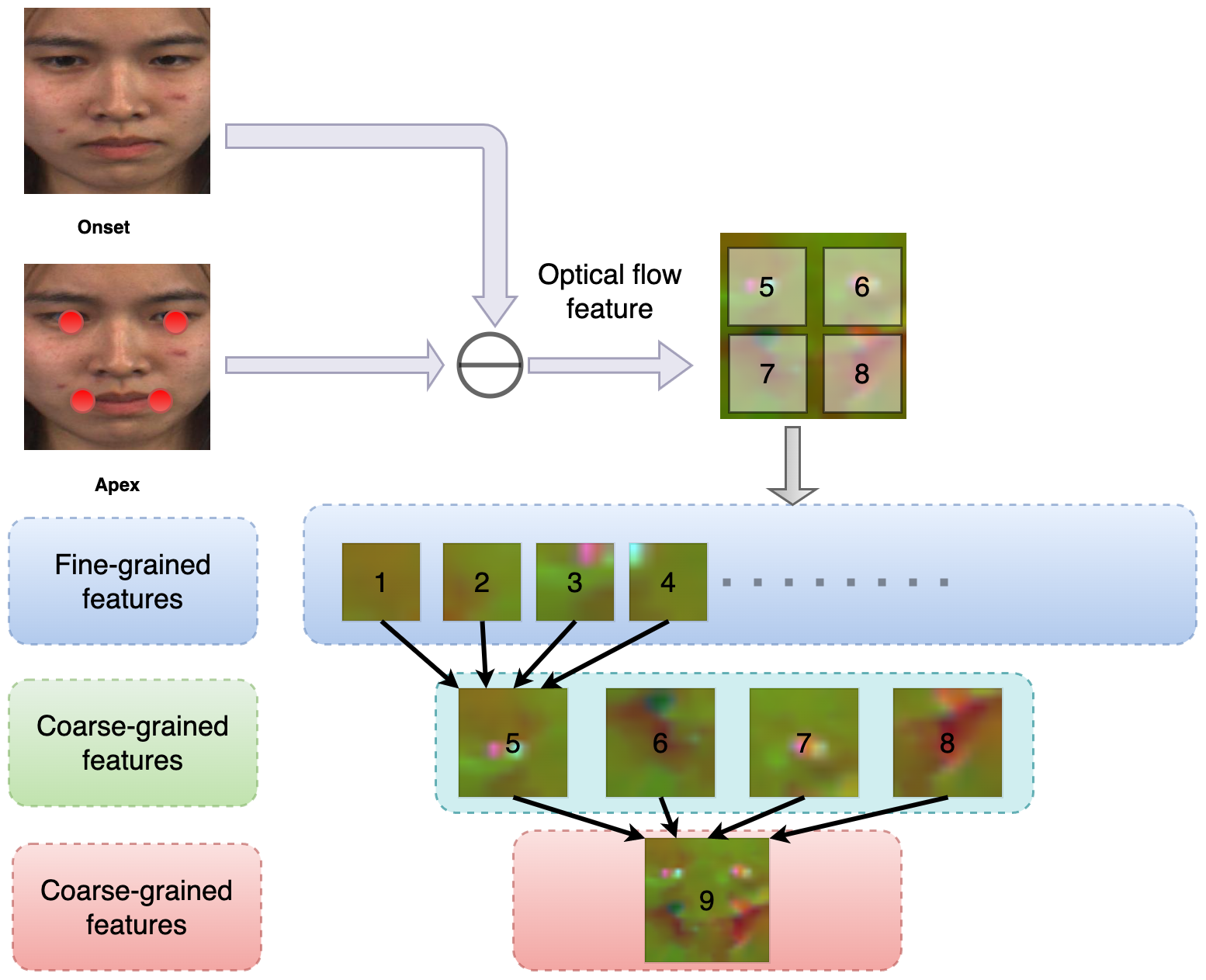}
  \caption{The proposed HTNet paradigm. The proposed framework focuses on four facial areas instead of entire facial regions-left eye area, right eye area, left lip area and right lip area, which can eliminate the audible background noise that the lab camera picked up because of potential light flicker. }
  \label{overall-view-structure}
\end{figure*}
In this study, we present a novel self-attention method utilizing Transformer layers to effectively capture both local and global interactions within a hierarchical structure. The low-level self-attention in the Transformer layers aims to capture fine-grained features within local regions. On the other hand, the high-level self-attention in these layers is designed to capture coarse-grained features spanning global regions. To facilitate interactions between different blocks at the same level, we propose an aggregation block. The overall architecture of our proposed method, referred to as HTNet, is illustrated in Fig. \ref{overall-view-structure}. Our contributions in this study can be summarized as follows:
\begin{itemize}
  \item [a)] 
   We introduce a novel self-attention mechanism of Transformer layers that effectively captures both local and global interactions within a hierarchical structure for recognizing micro-expressions in images. This is achieved by utilizing the proposed block aggregation function. The low-level self-attention focuses on capturing fine-grained features in local regions, while the high-level self-attention targets coarse-grained features in global regions.
  \item [b)]
 Our network specifically focuses on four facial areas - the left eye region, left lip region, right eye region, and right lip region - instead of considering the entire facial region. This approach helps to mitigate the impact of noticeable background noise that may be picked up by the lab camera due to possible light flickering. The use of Transformer layers allows us to concentrate on modeling small-scale, subtle muscle motions within each area through local self-attention. Additionally, the aggregation layer learns the interactions between the eye areas and lip areas. We conduct experiments to investigate how varying block sizes affect the accuracy of micro-expression recognition.
  \item [c)]
  Through experiments conducted on four available datasets, we demonstrate that our proposed method outperforms previous approaches noticeably. This highlights the effectiveness and superiority of our model for micro-expression recognition tasks.
\end{itemize}
\section{Related work}
\subsection{Conventional Methods}
In conventional techniques for expression recognition, appearance-based characteristics are commonly utilized. One prevalent pattern is the local binary pattern from three orthogonal planes (LBP-TOP) \cite{ref-20}. In some LBP-TOP works, LBP-TOP is transferred to the tensor-independent RGB space, which enhances robustness \cite{ref-21}. To achieve lower computational complexity, LBP-SIP effectively minimizes redundancy in LBP-TOP patterns, providing a lightweight representation \cite{ref-23}. LSDF \cite{ref-19} utilizes the 16 interest regions (ROIs) extracted by the Facial Action Coding System for micro-expression recognition. These 16 Regions of Interest represent 16 muscle motion areas, while irrelevant areas are eliminated. STCLQP \cite{ref-28} incorporates additional features such as orientation, shape attributes, and magnitude. By integrating sign, magnitude, and orientation components, STCLQP expands from LSDF and reduces the number of interest zones from 16 to 3. A codebook is created to extract discriminative and salient features from codebooks for different facial emotions.\\
In traditional methods, geometric-based features are extracted using facial landmarks or optical flow. These geometric-based features can identify motion deformations. Li \textit{et al.} \cite{ref-29} explore a deep learning technique for localizing face landmarks and dividing the facial area into interest zones. Since facial micro-expressions are produced by facial muscular contractions, assessing the orientation of these contractions is crucial for identifying emotions. Using this technique, they categorize the face into different areas of concern that correlate to different muscle action patterns. Furthermore, the facial area's activity can be adequately reflected by the optical flow. Liu \textit{et al.} \cite{ref-26} propose the use of a powerful MDMO feature extraction network for micro-expression identification. They calculate the optical flow for each image in the video sequence and divide the facial zones into various intriguing portions. Mean optical flow features are then calculated from each image, and an SVM classifier is employed on these mean optical flow features for emotion identification. Their method effectively considers both geographical position and regional statistical movement information, proving to be straightforward and efficient. Rather than using the LBP histogram, Liong \textit{et al.} \cite{ref-24} suggest weighting the histogram of LBP and averaging the value of optical flow features for recognizing emotions on the face, namely Bi-WOOF.\\
\subsection{Deep Learning Methods}
In recent years, an array of deep learning techniques has emerged for extracting facial characteristics in expression identification \cite{ref-33, zhang2015facial, zhang2017facial, niu2022four}. Xia \textit{et al.} \cite{ref-32} proposed the application of recurrent convolutional networks to establish connections between facial position information and record facial muscle contractions across various regions. This model incorporates multiple recurrent convolutional layers and a classification layer to capture visual characteristics for facial emotion recognition. Meanwhile, Lei \textit{et al.} \cite{ref-33} introduced the use of Transformer as an encoder to model the connections between nodes and edges of the face, constructing a face graph based on the edges and nodes of facial features. Although prior methods often relied on handcrafted features to represent subtle facial contractions, such features often come with higher computational costs. In response, Gan \textit{et al.} \cite{ref-9} proposed an automated approach to localize the apex frame and employ optical flow as input for their OFF-ApexNet. Leveraging CNN, their network extracts new feature descriptors from optical flow. On the other hand, some researchers have explored the adoption of shallow CNNs, which effectively derive high-level visual attributes from three components of movement estimation - horizontal, vertical optical flow fields, and optical strain - for inferring emotional states \cite{ref-5}. To address the issue of composite-database domain shift, Xia \textit{et al.} \cite{ref-6} developed an RCN model to examine how a smaller model affects the detection of micro-expressions. Within the RCN, they designed three parameter-free modules, including an attention unit, shortcut connection, and wide expansion, to prevent an increase in the number of learnable parameters. Additionally, researchers have acknowledged the significance of hierarchical structures in this context \cite{ref-45}.\\
\subsection{Transformer in Computer Vision}
The standard attention mechanism, initially introduced in the Transformer model, is known as scaled dot-product attention, which has become a fundamental method in Natural Language Processing (NLP) tasks. The Vision Transformer (ViT) \cite{ref-36} extends the Transformer architecture from NLP tasks to computer vision tasks. Specifically, for image classification, ViT applies the Transformer to non-overlapping image blocks. However, a significant drawback of ViT is its reliance on large training datasets to improve network performance. Recently, new training methodologies introduced by DeiT have allowed ViT to perform effectively with lower training datasets. Nevertheless, ViT is not well-suited as a backbone network for dense sub-vision tasks due to the quadratic increase in computing complexity with the increasing image size.To address this limitation and apply the Transformer architecture to high-resolution images, Liu \textit{et al.} \cite{ref-35} proposed the Swin Transformer. The Swin Transformer achieves this by employing shifted windows, which restrict self-attention calculation to a local window while enabling cross-window communication between different image blocks. This approach enhances the model's efficiency and scalability when dealing with high-resolution images. Other related Transformer works \cite{ref-37, ref-46} explore convolutional-free and simpler backbone networks for deep prediction tasks, offering alternative approaches to address the computational complexity and resource requirements in vision tasks.\\
\section{Proposed Method}
\begin{figure*}[t]
\centering
  \includegraphics[width=0.8\linewidth]{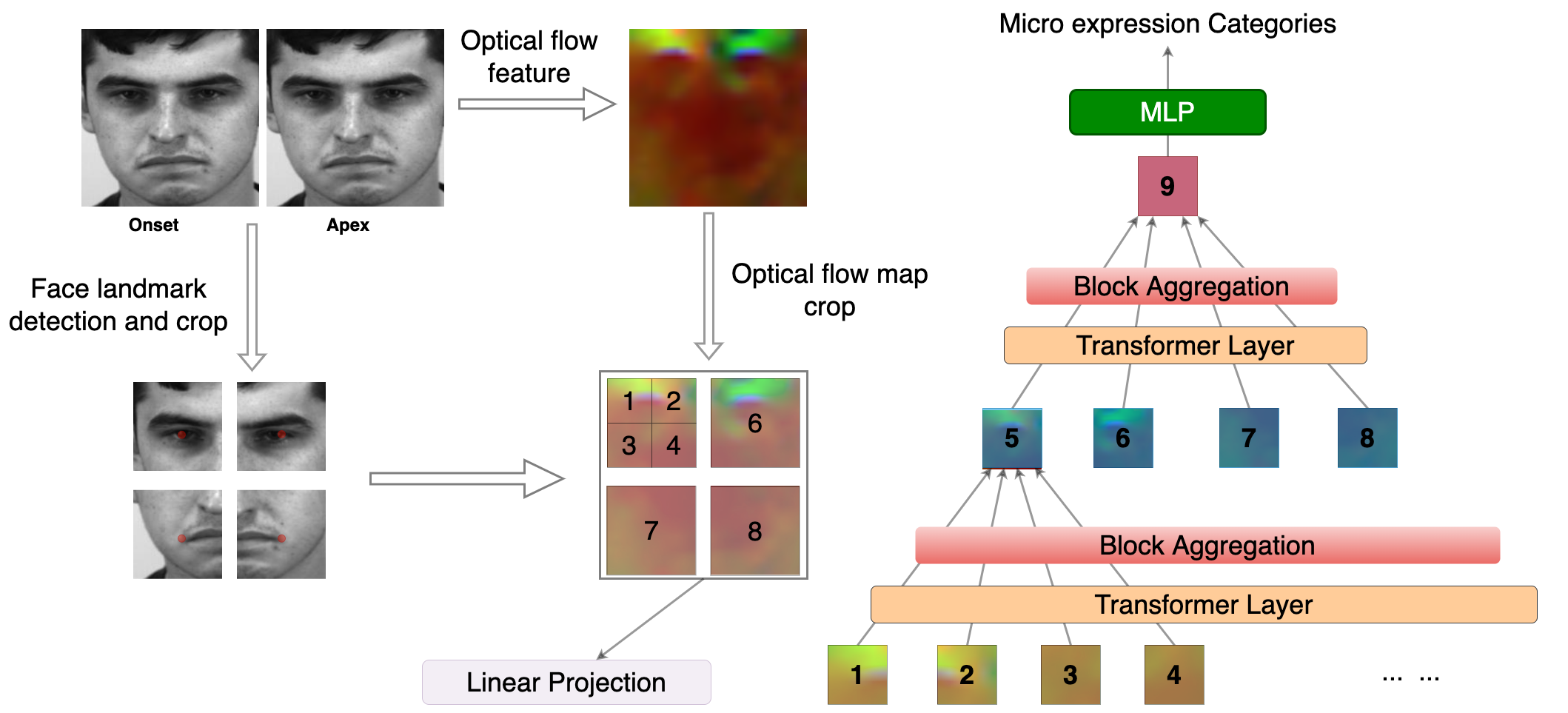}
  \caption{HTNet: Overall architectures of hierarchical transformer network for micro-expression recognition.  Low-level self-attention in transformer layers captures fine-grained features in local regions. High-level self-attention in transformer layers captures coarse-grained features in global regions. An aggregation block is proposed to create interactions between different blocks at the same level.}
  \label{fig-network}
\end{figure*}

As illustrated in Fig. \ref{fig-network}, the proposed HTNet consists of transformer layers and block aggregation at each level of the hierarchical structure. The transformer layer performs self-attention on each image block independently. At the low-level network, the self-attention function in the transformer layer captures fine-grained features. Subsequently, the block aggregation process aggregates small image blocks into larger ones, allowing for the creation of interactions between different blocks at the same level. This aggregation leads to the capture of coarse-grained features after each block aggregation. It is important to note that all the blocks within the same level share the same set of parameters. Finally, the MLP block in our model is applied to the final feature maps for micro-expression classification. This modular and hierarchical design enables HTNet to effectively extract and integrate features at different scales and levels of granularity for improved micro-expression recognition performance.
\subsection{Optical Flow Map Extraction}
Optical flows, generated from the commencement and peak frames, serve as a valuable means to describe movement displacements in facial areas. These optical flows have demonstrated promising results in micro-expression recognition datasets \cite{ref-9,ref-5}. \\
In order to obtain an optical feature map, it is essential to determine the onset and apex frame indices. However, in the case of micro-expression datasets, the onset frame index is already provided, necessitating the determination of only the apex frame index from video sequences. To achieve this, we adopt the D\&C-RoIs method, which has been widely used in prior research on micro-expression \cite{ref-9}. The D\&C-RoIs approach effectively establishes the relationship between the commencement frame and subsequent frames, allowing for the accurate identification of the apex frame index. This ensures the reliable extraction of optical flow features for subsequent micro-expression recognition tasks.
\begin{equation}
\begin{aligned}
d =\frac{\sum_{i=1}^{B} h_{1i} \times h_{h2i}}{\sqrt{\sum_{i=1}^{B} h_{1i}^2 \times \sum_{i=2}^{B} h_{2i}^2 }}
\end{aligned}
\label{apex-spotting}
\end{equation}
,where $B$ is the number of bins in histograms, $h_1$ is the first frame, and $h_2$ is the other frames except the first frame. The highest rate of difference in ROIs will be selected, which can represent the apex frame that has maximum facial muscle changes. \\
Then, we obtain optical flow feature maps using the commencement and peak frames. It is possible to formulate the optical flow feature map as follows:
\begin{equation}
\begin{aligned}
V ={(u(x, y), v(x,y)) | x = 1,2,....., X, y = 1,2, ..., Y}
\end{aligned}
\label{optical-u-v}
\end{equation}
, where X and Y represent the frame's width $W$ and height $H$ respectively, $u(x,y)$ and  $v(x,y)$  is the horizontal  and  vertical component of optical flow feature map $V$, $V = [V_x, V_y]$, where $V \in R^{W\times H \times 2}$.\\
In our approach, we employ the first-order derivatives of the optical flow field to calculate the variations in the optical flow fields, commonly known as optical strain. The optical strain provides an estimation of the degree of facial displacement, thereby offering valuable insights into the subtle movements that occur during micro-expressions. This computation of optical strain allows us to capture and analyze the intricate facial dynamics, contributing to the accurate recognition of micro-expressions:
\begin{equation}
\begin{aligned}
V_z =\sqrt{  \frac{\partial V_x}{\partial x}^2  + \frac{\partial V_y}{\partial y}^2 +\frac{1}{2} ( \frac{\partial V_x}{\partial y}^2 + \frac{\partial V_y}{\partial x}^2   )  }
\end{aligned}
\label{variation-optical}
\end{equation}
,where $ \frac{\partial V_x}{\partial x}^2 $, $ \frac{\partial V_y}{\partial y}^2$, $\frac{\partial V_x}{\partial y}^2 $ and $\frac{\partial V_y}{\partial x}^2 $ are the  partial first-order derivatives of $V$. Finally, three-dimension optical flow feature maps are formed and represented as $V_m = [V_x,V_y,V_z]$ and  $V_m \in R^{W\times H \times 3}$. \\
In our network, we adopt a region-specific approach by focusing on four specific facial areas - the left eye region, left lip region, right eye region, and right lip region - instead of considering the entire facial region. This strategic selection aims to mitigate the impact of audible background noise that may be present in the lab recordings due to potential flickering lights. To extract the facial optical flow feature maps from the entire optical flow feature maps $V_m$, we employ the Multi-task Cascaded Convolutional Networks (MTCNN) \cite{ref-42} to obtain the face landmark coordinates from the apex images. The four facial optical flow feature maps are subsequently cropped from $V_m$, representing the left-eye optical flow feature map, left-lip visual feature map, right-eye optical flow feature map, and right-lip feature map. Each of these four feature maps has a size of $\frac{W}{2} \times \frac{H}{2} \times 3$, which is half the size of the whole optical flow image. Following the extraction of the four optical flow features, we combine them and feed the combined features into our HTNet for micro-expression recognition. The entire process is visually depicted in Fig. \ref{fig-network}.\\
\subsection{Transformer layer}
In our approach, each image block has a size of $P \times P$ when processing an input optical flow image with dimensions $H \times W \times 3$. After linear projection and partitioning on the optical flow images, each patch has a feature dimension of $P \times P \times 3$. Subsequently, the patches are flattened, resulting in an input for our model denoted as $X \in \mathbb{R}^{b \times H_n \times n \times d}$, where $H_n$ represents the number of blocks in each level, $b$ is the batch size, $n$ is the sequence length, and $H_n \times n = \frac{H \times W}{P^2}$. Multiple transformer layers are applied within each block, and the hierarchy determines the number of transformer layers utilized. Each transformer layer consists of Layer Normalization (LN), Multi-Head Self-Attention (MSA) layer, and Feed-Forward Fully Connected Network (FFN). To encode spatial information, a trainable positional embedding vector is incorporated into all sequence vectors in $\mathbb{R}^d$. This ensures that spatial relationships and positional information are effectively captured and encoded within the feature representations:
\begin{equation}
\begin{aligned}
Y_l^{'i} &= Y_{l-1}^{i}  + MSA(LN(Y_{l-1}^{i} ))\\
Y_l^{i+1} &= Y_{l}^{'i}  + FFN(LN(Y_{l}^{'i} ))\\
Y_l^{'i+1} &= Y_{l}^{i+1}  + MSA(LN(Y_{l}^{i+1} ))\\
Y_l^{i+2} &= Y_{l}^{'i+1}  + FFN(LN(Y_{l}^{'i+1} ))\\
\end{aligned}
\label{transformer-layer-eqation1}
\end{equation}
,where $l= 1,2,..., L$, $l$ is the index of the $l-th$ block in each hierarchy layer $i$, and $L$ is the overall number of blocks in each hierarchy layer. \\
\begin{figure*}[t]
\centering
  \includegraphics[width=0.4\linewidth]{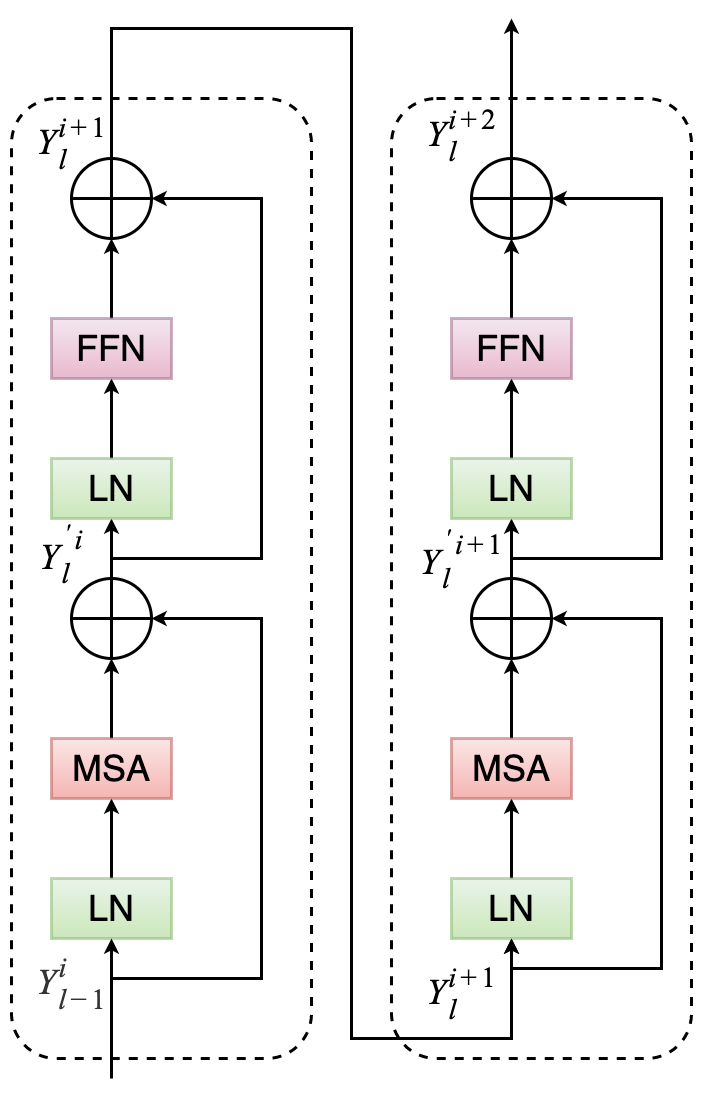}
  \caption{Several transformer layers will be applied in each block in parallel.Hierarchy determines the amount of transformer layers. The transformer layers is composed of Layer normalization (LN), multi-head self-attention (MSA) layer and feed-forward fully connected network (FFN). The spatial information will be encoded by adding a trainable positional embedding vector to all sequence vectors in $R^d$.}
  \label{transformer-layer-structure}
\end{figure*}
The FFN includes two layers: $max(0, xW_1 + b)W_2  + b$. At each block $i$ within the same level, the multi-head self-attention mechanism is applied. In this self-attention component, the input $X \in \mathbb{R}^{n \times d}$ is transformed into three parts, namely queries $Q$, keys $K$, and values $V$, where $n$ represents the sequence length and $d$ is the dimension of the inputs. Subsequently, the scaled dot-product attention is employed on $Q$, $K$, and $V$:
\begin{equation}
\begin{aligned}
MSA(Q,K,V) = softmax(\frac{QK^T}{\sqrt{d}})V.
\end{aligned}
\label{self-attention-eqation1}
\end{equation}
LN will be applied in each block as follows:
\begin{equation}
\begin{aligned}
LN(x) =\frac{x-\mu}{\delta}o \lambda +\beta
\end{aligned}
\label{normalization-eqation1}
\end{equation}
,where $\mu$ is the mean of features and $\delta$ is standard deviation of the feature, $o$ is the element-wise dot and $ \lambda$ and $ \beta$ are learnable parameters. \\
Following the transformer layer, we employ block aggregation to merge the output of the transformer layer. Specifically, we group every four small blocks into one larger block through the block aggregation process. 
\subsection{Block Aggregation}
The block aggregation function utilized in our HTNet shares similarities with several Pyramid designs. However, a notable distinction is that our model employs local attention on each image block, rather than global attention on the entire image. This approach proves to be beneficial for enhancing the model's performance since micro-expression recognition heavily relies on localized facial muscle motion areas. By focusing on specific face regions and leveraging local attention, our model effectively captures the essential features for inferring micro-expression states while disregarding irrelevant facial areas.\\
\begin{figure*}[t]
\centering
  \includegraphics[width=0.6\linewidth]{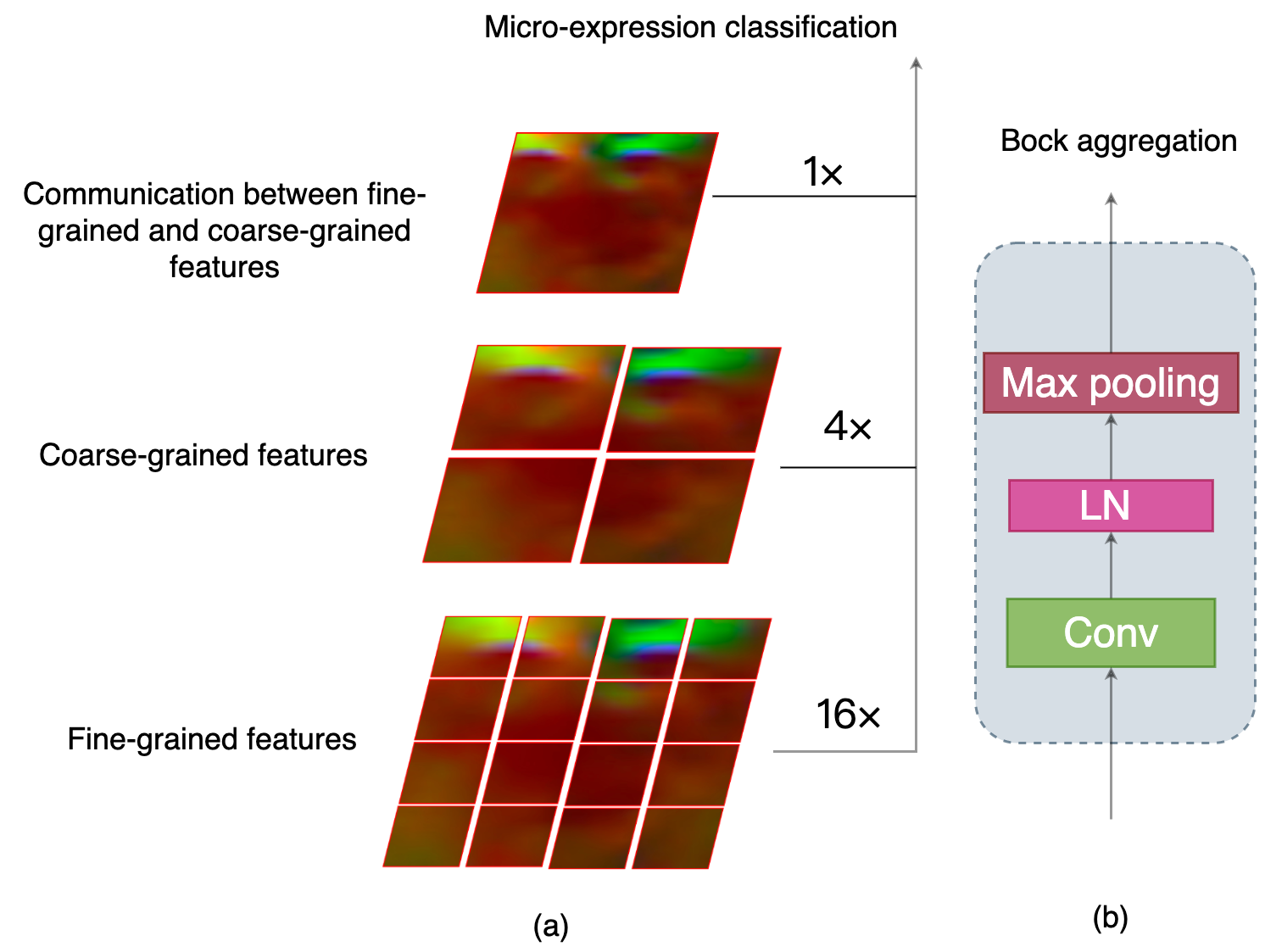}
  \caption{The block aggregation include $3 \times 3$ convolutional layer and followed by LN and $3 \times 3$ max pooling. At the bottom of our model, the facial optical flow map includes 16 facial blocks, which are $4 \times 4$ feature maps. By using  $3 \times 3$ convolutional layer on this feature map, partial feature maps within four facial areas will be merged and the block size will become $2 \times 2$ blocks which corresponding to four facial areas-left eye region, left lip region, right eye region and right lip region.}
  \label{block-aggregation}
\end{figure*}\\
In our HTNet model, each block independently processes the optical flow map, and block communication occurs during block aggregation. Block aggregation operates on four neighboring blocks, facilitating the exchange of information between them and enabling the fusion of local and global features. Specifically, low-level block aggregation focuses on exchanging information within the four facial areas - left eye region, right eye region, left lip region, and right lip region. This process extracts fine-grained features that capture detailed facial dynamics. On the other hand, high-level block aggregation facilitates global information exchange between the four facial areas, resulting in the extraction of coarse-grained features that capture broader facial expressions. The overall process is depicted in Fig. \ref{block-aggregation}. At hierarchy $l$, the optical flow image size is represented as $X_l \in \mathbb{R}^{b \times h \times w \times d^{l}}$. After block aggregation, the optical flow image size becomes $X_{l+1} \in \mathbb{R}^{b \times 2h \times 2w \times d^{l+1}}$. Subsequently, all four facial areas are merged into one facial feature map, denoted as $X \in \mathbb{R}^{b \times H \times W \times D}$, with $d^{l+1} > d^{l}$ to preserve and enhance the features effectively.\\
The block aggregation process in HTNet consists of a $3 \times 3$ convolutional layer, followed by Layer Normalization (LN) and $3 \times 3$ max pooling. At the bottom of our model, the facial optical flow map consists of 16 facial blocks, each represented by a $4 \times 4$ feature map. The first $3 \times 3$ convolutional layer merges partial feature maps within the four facial areas, reducing the block size to $2 \times 2$, corresponding to the left eye area, left lip region, right eye area, and right lip region. Subsequently, another $3 \times 3$ convolutional layer facilitates information exchange between the four facial areas, further reducing the block size to $1 \times 1$, and extracting full optical feature maps. Finally, the extracted full feature maps are fed into the MLP layer for micro-expression classification. This hierarchical process effectively captures and integrates features at different levels of granularity, contributing to improved micro-expression recognition.\\
\subsection{Loss function}
In this paper, we employ the cross-entropy loss function to train our model. The computation of the cross-entropy loss $L$ can be expressed by the following formula:
 \begin{equation}
\begin{aligned}
L = \sum_i(-w_i log(p_t^i))\\
p_t = p_c^y(1-p_c)^{1-y}
\end{aligned}
\label{loss-entropy}
\end{equation}
,where $w_i$ is the weight of samples in the dataset, y is the label and $y_i \in {0,1}$.
\\
\section{Experiments}
\subsection{Datasets}
The experiments are conducted on four databases:  SAMM\cite{ref-48}, SMIC\cite{ref-49}, CASME II\cite{ref-50} and CASME III \cite{ref-51} databases. To ensure consistency and comparability, SAMM, SMIC, and CASME II are merged into a composite dataset, where the same emotion labels from these three datasets are adopted for the micro-expression recognition tasks.  In these datasets, the emotion categories are divided as follows: the ``positive'' emotion category includes the ``happiness'' emotion class, and the ``negative'' emotion category includes ``sadness'',``disgust'', ``contempt'',  ``fear'' and ``anger'' emotion classes while ``surprise'' emotion category only includes ``surprise'' class. \\
\begin{table*}[!ht]
\caption{ The experiments are implemented on SAMM\cite{ref-48}, SMIC\cite{ref-49}, CASME II\cite{ref-50} and CASME III \cite{ref-51} databases. SAMM, SMIC, and CASME II are merged into one composite dataset, and the same labels in these three datasets are adopted for micro-expression tasks. 
}
 \label{datasets-compare}
\centering
\begin{tabular}{c c c c c c} 
\hline \hline
 Database &SAMM& CASME II  & SMIC  & CASME III  \\ 
\hline
Subjects & 28&  24 & 16 &  100\\ 
\hline
 Samples &133 & 145  & 164 & 943  \\ 
\hline
 Frame rate &200 & 200 &  100& 30 \\ 
\hline
 Cropped image resolution &$28 \times 28$ & $28 \times 28$ & $28 \times 28$ & $28 \times 28$\\ 
\hline
Negative & 92 & 88 & 70 & 508 \\
Positive & 26 & 32 & 51 & 64 \\
Surprise & 15 & 25 & 43 & 201 \\
\hline
Onset index & \checkmark & \checkmark & \checkmark & \checkmark\\
Offset index & \checkmark & \checkmark & \checkmark &\checkmark \\
Apex index & \checkmark & \checkmark &    $\times$ &\checkmark  \\
\hline
\end{tabular}
\end{table*}

\textbf{SAMM} \cite{ref-48}: The SAMM dataset comprises 28 participants, 133 micro-expressions, and 147 long videos featuring 343 macro-expressions. The dataset is rich in Action Units coding, providing comprehensive facial expression information. The onset, offset, and apex indices for micro-expressions are also provided in SAMM. The original samples in the dataset have a resolution of 2040 by 1088 pixels, and the frame rate is set at 200 frames per second. To facilitate the experiment, we cropped the face images from the original samples, resulting in face images with a resolution of $28 \times 28$ pixels. Emotion categories in images in SAMM are divided into categories including ``disgust'', ``fear'', ``contempt'', ``angry'', ``repression'', ``surprise'', ``happiness'', and ``others''. After classifying into three emotion categories, the number of ``negative'', ``positive'', and ``surprise'' is 92, 26,15. 

\textbf{CASME II} \cite{ref-50}: The CASME II dataset consists of data from 24 subjects, totaling 145 samples corresponding to 145 emotions. All samples are captured using lab cameras, with a frame rate set at 200 frames per second. The original size of the samples is $640 \times 480$ pixels. For our experiment, we cropped and resized the face images to a resolution of $28 \times 28$ pixels. Samples in CASME II are divided into categories such as ``happiness'', ``surprise'', ``disgust'', ``sadness'', ``fear'', ``repression'', and ``others''. After merging into three emotion categories, the number of ``negative'', ``positive'', and ``surprise'' is 88, 32,25. The onset, offset, and apex index are annotated in CASME II. 

\textbf{CASME III} \cite{ref-51}:CASME III, the third generation of the Facial Spontaneous Micro-Expression database, is distinguished by its inclusion of depth information and high ecological validity, making it a valuable resource for micro-expression recognition. CASME III Part A comprises data from 100 subjects, totaling 943 samples corresponding to 943 emotions. The samples are captured using a lab camera, with a frame rate of 30 frames per second, and have an original resolution of $1280 \times 720$ pixels. Samples in CASME III part A are categorized into ``happiness'',  ``anger'',``fear'',``disgust'',   ``surprise'', ``others'' and ``sadness''. The total number of ``negative'', ``positive'', and ``surprise'' is 508, 64, and 201. 

\textbf{SMIC}\cite{ref-49}: The SMIC-HS dataset comprises data from 16 subjects, totaling 164 samples corresponding to 164 emotions. All samples are captured using a lab camera with a frame rate of 100 frames per second. The original image size of the samples is $640 \times 480$ pixels. To focus on the facial region of interest and maintain a consistent input size for our micro-expression recognition task, we crop the face images to a resolution of $28 \times 28$ pixels. Samples in SMIC are categorized into ``negative'', ``surprise'' and ``positive''. The number of ``negative'', ``positive'', and ``surprise'' is 70, 51,43. The onset and offset are given in SMIC, but the apex index is not given in SMIC. The detailed information of these three datasets can be summarised in Table \ref{datasets-compare}

\subsection{Implementation Details}
Initially, in the case of the SMIC dataset, where the peak frame index is missing, we employed the D\&C-RoIs technique \cite{ref-39} to determine the index of the peak frame. For the SAMM, CASME II, and CASME III datasets, the ground truth for the peak frame is available, which simplifies the process of obtaining the crucial micro-expression moment. After obtaining the onset and apex images from the dataset, we utilized Gunnar Farneback's algorithm \cite{ref-41} to extract optical flow from these images at both the commencement and peak time points.  Next, the three elements of optical flow images - horizontal, vertical, and optical strain elements - are resized to $28 \times 28 \times 3$ pixels. Afterward, we employ the Multi-task Cascaded Convolutional Networks (MTCNN) \cite{ref-42} to extract face landmark coordinates from the apex images. These face landmark coordinates are vital for localizing specific facial regions with high accuracy. Based on the face landmark coordinates, we extract four important face areas, namely the left-eye, left-lip, right-eye, and right-lip optical flow feature maps. These four optical flow feature maps are half the size of the entire optical flow image, measuring $14 \times 14 \times 3$ pixels. By focusing on these specific facial regions, we can effectively capture the relevant facial muscle movements associated with micro-expressions.Following the extraction of the four optical flow feature maps, our HTNet network receives and combines them. This comprehensive approach ensures that our HTNet is well-equipped to recognize and classify micro-expressions accurately and efficiently.

The experiments are conducted using PyTorch and Python 3.9 on Ubuntu 22.04 operating system. We set the learning rate for the training parameters to $5 \times 10^{-5}$ and use a maximum of 800 epochs for training. 

\textbf{Setup}: The standard evaluation method for micro-expression tasks is the leave-one-subject-out (LOSO) cross-validation. LOSO cross-validation is preferred as it allows for a fair comparison between different models and ensures that the model's performance is not biased by specific subject characteristics. This approach closely simulates real-world scenarios, where individuals with diverse backgrounds and expressions are encountered in various settings and locations.

\textbf{Performance metrics}
The class distribution in the composite micro-expression dataset is unbalanced, with different emotions having varying frequencies. Specifically, the rate of surprise, positive, and negative emotions is approximately 1:1.3:3, respectively.To address this issue, we employ the Unweighted Average Recall (UAR) and Unweighted F1 score (UF1) to report our results. 

\begin{itemize}
  \item [1) ] The Unweighted F1-score (UF1), also known as the macro-average F1-score, is a metric that is commonly used to evaluate performance in multi-class classification tasks with imbalanced class distributions. To calculate the UF1, we need to compute the False Positives (FP), True Positives (TP), and False Negatives (FN) for each class $c$ in all folds of the leave-one-subject-out (LOSO) cross-validation. Then, the F1-score for each class $F1_c$ can be computed using the formula:
\begin{equation}
\begin{aligned}
F1_c &=\frac{2\times TP_c}{2\times TP_c +FP_c +FN_c}\\
UF1 &=\frac{F1_c}{C}\\
\end{aligned}
\label{uf1}
\end{equation}
In the formula \ref{uf1}, the C is the number of classes.
  
  \item [2) ] Unweighted Average Recall (UAR): UAR is a metric that is particularly useful when evaluating the effectiveness of a model in the presence of imbalanced class ratios. To calculate the UAR, we first compute the True Positives ($TP_c$) for each emotion class $c$. The $TP_c$ represents the number of correctly classified samples in each class. Additionally, we calculate the total number of samples ($n_c$) in each class. The number of emotion classes is denoted as $C$.
 \begin{equation}
\begin{aligned}
UAR = \frac{1}{C}\sum_{c}{\frac{TP_c}{n_c}}
\end{aligned}
\label{uar}
\end{equation}
\end{itemize}

\subsection {Comparison to State of the Arts}
In Table \ref{sota-results-composite} and Table \ref{sota-results-part-A}, we present the performance of our proposed method in comparison with previous handcrafted and deep learning methods on the micro-expression datasets, namely CASME II, CASME III, SMIC, and SAMM. The evaluation metrics used are Unweighted F1-score (UF1) and Unweighted Average Recall (UAR). The results are reported.Bold text is used to highlight the best result achieved among the methods for each dataset and metric.

LBP-TOP \cite{ref-23}:  The authors introduced a novel approach for micro-expression recognition by utilizing local binary patterns with six intersection points (LBP-SIP) to extract facial features. The LBP-SIP method effectively reduces redundancy in LBP-TOP patterns, resulting in reduced computational complexity. Additionally, they applied a Gaussian multi-resolution pyramid to extract features at different image resolutions and then concatenated these features for micro-expression recognition. 

Bi-WOOF \cite{ref-24}: To extract essential facial features from the apex image, they introduced the Bi-Weighted Optical Flow method. This technique effectively captures the relevant facial movements and highlights the important information needed for micro-expression recognition.

OFF-ApexNet \cite{ref-9}: Because implementing optimal feature extraction methods for the subtle motion of facial expressions is complex, and many approaches extract the features of the subtle motion of facial expressions by using handcraft features.For micro-expression tasks, they suggested utilising optical flow fields from the beginning and peak frames. In the optical flow fields, horizontal and vertical features are obtained and fed into CNN-based network for further feature enhancement. After that, the extracted features from OFF-ApexNet will be used for micro-expression classification. 
\begin{table*}[!t]
\centering
\scriptsize
\caption{ The Unweighted F1-score (UF1) and Unweighted Average Recall (UAR) performance of handcraft methods, deep learning methods and our HTNet method under LOSO protocol on the composite (Full), SMIC, CASME II and SAMM. Bold text indicates the best result.}
\begin{tabular}{c c c c c c c c c c c c c c c  c }
\hline
 \multirow{2}{*}{Approaches} &\multicolumn{3}{c}{Full} &\multicolumn{3}{c}{SMIC}&\multicolumn{3}{c}{CASME II}&\multicolumn{3}{c}{SAMM}\\
\cline{2-3} \cline{5-6} \cline{8-9} \cline{11-12}
 &UF1 &UAR& &UF1 &UAR &  &UF1 &UAR &  &UF1 &UAR& \\
\hline
LBP-TOP \cite{ref-23} &0.5882 &0.5785& &0.2000 &0.5280 &  &0.7026 &0.7429 &  &0.3954 &0.4102 &\\
Bi-WOOF \cite{ref-24}&0.6296 &0.6227& &0.5727 &0.5829 &  &0.7805 &0.8026 &  &0.5211 &0.5139& \\
\hline
AlexNet \cite{ref-11} &0.6933 & 0.7154 & &0.6201& 0.6373 & &0.7994 & 0.8312 & &0.6104  &0.6642& \\
GoogLeNet \cite{ref-13}    &0.5573& 0.6049& & 0.5123& 0.5511& & 0.5989 & 0.6414 & &0.5124 & 0.5992& \\
VGG16 \cite{ref-14}          &0.6425 &0.6516 & & 0.5800 &0.5964 & &0.8166 & 0.8202  & &0.4870 & 0.4793& \\
OFF-ApexNet \cite{ref-9}&0.7196 & 0.7096 & & 0.6817 & 0.6695 & & 0.8764 & 0.8681& & 0.5409 &0.5392& \\
STSTNet \cite{ref-5}     &0.7353  & 0.7605  & &0.6801& 0.7013 & &0.8382 & 0.8686& & 0.6588 &0.6810& \\
CapsuleNet \cite{ref-43} &0.6520 & 0.6506 & & 0.5820 & 0.5877 & & 0.7068 &0.7018 & & 0.6209& 0.5989& \\
Dual-Inception \cite{ref-44}&0.7322 &0.7278& & 0.6645 & 0.6726 & & 0.8621 & 0.8560 & &0.5868 & 0.5663& \\
EMR  \cite{ref-8}              &0.7885 &0.7824  & &\underline{0.7461}& \underline{0.7530} & & 0.8293 &0.8209 & &\underline{0.7754} & 0.7152& \\
RCN  \cite{ref-6}   &0.7432 &0.7190& &0.6326 &0.6441 &  &0.8512 &0.8123 &  &0.7601 &0.6715& \\
FeatRef \cite{ref-15} &\underline{0.7838} & \underline{0.7832}  & &0.7011 & 0.7083 & & \underline{0.8915} & \underline{0.8873} & &0.7372 & \underline{0.7155}& \\
HTNet(Ours) &\textbf{0.8603} & \textbf{0.8475}  & &\textbf{0.8049} & \textbf{0.7905} & & \textbf{0.9532} & \textbf{0.9516} & &\textbf{0.8131} & \textbf{0.8124}& \\
\hline
\end{tabular}
 \label{sota-results-composite}
\end{table*}

STSTNet \cite{ref-5}: The authors proposed employing a three shallow CNN-based model to obtain high-level discriminative representations for classifying micro-expression emotions. The CNN-based network consists of several layers, including convolutional layers, fully connected layers, and pooling layers. These layers work together to extract relevant muscle motion features from two directions: the horizontal and vertical optical flow fields, as well as optical strain information. 

Dual-Inception \cite{ref-44}: To address the challenges of cross-database micro-expression recognition, the authors proposed a novel approach using two inception networks to extract horizontal and vertical features from the optical flow maps. By feeding the horizontal component of the optical flow into one inception network and the vertical component into another inception network, they can independently capture relevant patterns and information from both directions.

EMR  \cite{ref-8}: Given the finite number of training samples in facial emotion datasets, it becomes essential to enhance both the quantity and quality of the available training images. To address this, the proposed method in EMR incorporates two domain adaptation strategies.The first strategy involves adversarial training methods.The second strategy is the expression magnification method

RCN  \cite{ref-6}: In the context of the composite dataset, the subtle facial motion necessary for micro-expression recognition may be lost due to domain shift, leading to a decline in the model's performance. To address this issue, the proposed approach suggests using a lower-size image as input and adopting a smaller-architecture model, which has shown to be beneficial for improving the model's performance on composite dataset tasks.

FeatRef \cite{ref-15}: In the FeatRef network, it includes two stages, in the first stage, horizontal inception network and vertical inception network will extract horizontal and vertical muscle motion features. After that, the  horizontal  and vertical features will be merged and fed into three attention based network for classifying these extracted features into different micro-expression categories. Finally, the classification branch is used to fuse salient and discriminative features obtained from the inception module for inferring micro-expression. Their experiments demonstrated that Feature Refinement (FeatRef) could extract discriminative and salient representation for micro-expression recognition and obtained good performance in micro-expression datasets. 

\subsubsection{Compared to Handcrafted  Methods}

Table \ref{sota-results-composite} presents a comparative analysis of different methods, including LBP-TOP and Bi-WOOF, which extract facial features using appearance-based and geometric-based techniques, respectively. Both methods employ SVM as the classifier. In contrast, our proposed method, HTNet, demonstrates substantial improvements in UF1 and UAR for composite datasets. Specifically, the UF1 increases from 0.6296 to 0.8603, and the UAR improves from 0.6227 to 0.8475, showcasing an improvement of more than 20\%. Furthermore, HTNet consistently outperforms handcrafted methods (LBP-TOP and Bi-WOOF) as well as various deep learning methods (e.g., GoogLeNet and VGG) on CASME II, SMIC, and SAMM datasets. These results underscore the efficacy of HTNet in addressing domain shift and achieving superior performance in micro-expression recognition tasks compared to both handcrafted and deep learning approaches.
\begin{table}[!t]
\centering
\small
\caption{ The Unweighted F1-score (UF1) and Unweighted Average Recall (UAR) performance of deep learning methods and our HTNet method under LOSO protocol on CASME III Part A. Bold text indicates the best result.}
\begin{tabular}{c c c c c c c c c c c   }
\hline
 \multirow{2}{*}{Approaches} &\multicolumn{3}{c}{CASME III Part A}\\
\cline{2-3} 
 &UF1 &UAR&  \\
\hline
AlexNet \cite{ref-11}  &0.257 & 0.2634 &  \\
STSTNet  \cite{ref-5}    &0.3795  & 0.3792  &  \\
RCN      \cite{ref-6}   &\underline{0.3928} &\underline{0.3893}& \\
FeatRef  \cite{ref-15}&0.3493 & 0.3413  & \\
HTNet(Ours) &\textbf{0.5767} & \textbf{0.5415}  &  \\
\hline
\end{tabular}
 \label{sota-results-part-A}
\end{table}
\subsubsection{Compared to Deep Learning Methods}
In Table \ref{sota-results-composite} and Table \ref{sota-results-part-A}, our HTNet outperforms most deep learning methods by a considerable margin. According to Table \ref{sota-results-composite}, HTNet achieved an UF1 and UAR of 0.8603 and 0.8475, respectively, on the full composite dataset, which represents an improvement of approximately 7\% compared to previous state-of-the-art methods. Analyzing Table \ref{sota-results-composite}, we observe that ALexNet, GoogLeNet, and VGG16 achieved UF1 and UAR of 0.6933 and 0.7154, 0.5573 and 0.6049, and 0.6425 and 0.6516, respectively, on the full composite dataset. These deep learning methods yielded inferior results compared to other deep learning methods in the full composite datasets. The reason for this outcome lies in the fact that these three methods use deeper networks, which may lead to learning noise information due to the limited number of training samples.In contrast, HTNet focuses on four facial areas instead of the entire facial region, effectively eliminating background noise picked up by the lab camera due to potential light flicker. With local self-attention in each face region, the transformer layer can concentrate on localizing subtle muscle contractions. Furthermore, the aggregation layer facilitates learning interactions between different resolutions of optical flow feature maps. As a result, HTNet achieves superior performance compared to previous methods on full composite datasets.

In Table \ref{sota-results-part-A}, we conducted experiments on CASME III Part A and reported the Unweighted F1-score (UF1) and Unweighted Average Recall (UAR) of deep learning methods, including our HTNet, under the leave-one-subject-out (LOSO) protocol. HTNet outperforms previous methods by a significant margin, demonstrating its effectiveness in micro-expression recognition.

\subsection {Ablation study}
This section presents an in-depth analysis of the impact of various parameters in our HTNet model. We investigate the influence of block size, hidden dimension, number of heads in the transformer, and the number of transformer layers at each hierarchy level. The initial experimental settings include a block size of $7\times 7$ at the bottom-level, a hidden dimension of $256$, three heads in the transformer, and $(2,2,8)$ transformer layers in each hierarchy level. For each ablation experiment, we modify a specific parameter while keeping other settings constant. The evaluation of each approach is conducted using the UAR metric and UF1 metric.

\subsubsection{Impacts of block size}
\begin{table*}[t]
\centering
\caption{ Study the impacts of different block sizes on composite datasets-SMIC, SAMM and CASME II. The composite datasets' Unweighted F1-score (UF1) and Unweighted Average Recall (UAR) performance are reported. }
\begin{tabular}{c| c c c c c c    }
\hline
\#Block size (Bottom level) &$5 \times 5$ &$6 \times 6$ &$7 \times 7$ &$8 \times 8$ &$9 \times 9$ & $10 \times 10$ \\
\#Block size (Middle level) &$10 \times 10$ &$12 \times 12$ &$14 \times 14$ &$16 \times 16$ &$18 \times 18$ & $20 \times 20$ \\
\#Block size (Top level) &$20 \times 20$ &$24 \times 24$ &$28 \times 28$ &$32 \times 32$ &$36 \times 36$ & $40 \times 40$ \\
\hline
Full UF1   &0.837 &0.8271  & \textbf{0.8603}   & 0.85 & 0.8511 & 0.8546\\
Full UAR  &0.8213 &0.8037  & 0.8475  & \textbf{0.846}   & 0.8445 & 0.8383 \\
SMIC UF1   &0.7556 &0.7394  & \textbf{0.8049}   & 0.7833 & 0.7753 & 0.8008\\
SMIC UAR  &0.7469 &0.7214  & \textbf{0.7905}  & 0.7792   & 0.7708 & 0.7892 \\
SAMM UF1   &\textbf{0.8237} &0.7903  & 0.8131   & 0.7909 & 0.8168 & 0.8068\\
SAMMUAR  &0.8033 &0.775  & \textbf{0.8124}  & 0.79   & 0.8088 & 0.7812 \\
CASME II UF1   &0.9422 &0.9482  & 0.9532   & \textbf{0.964} & 0.9722 & 0.957\\
CASME II UAR  &0.9308 &0.9317  & 0.9516  & 0.962   & \textbf{0.9658} & 0.945 \\
\hline
\hline
\end{tabular}
 \label{impacts_of_block_size}
\end{table*}

 \begin{figure}[!t]
\centering
\subfloat[]{\includegraphics[width=0.48\linewidth]{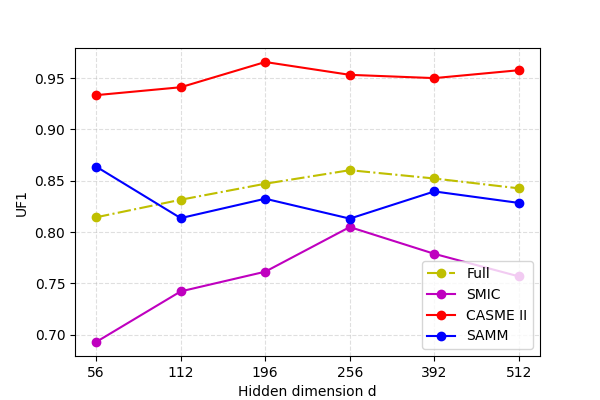}%
\label{dimension-fig_first_case}}
\subfloat[]{\includegraphics[width=0.48\linewidth]{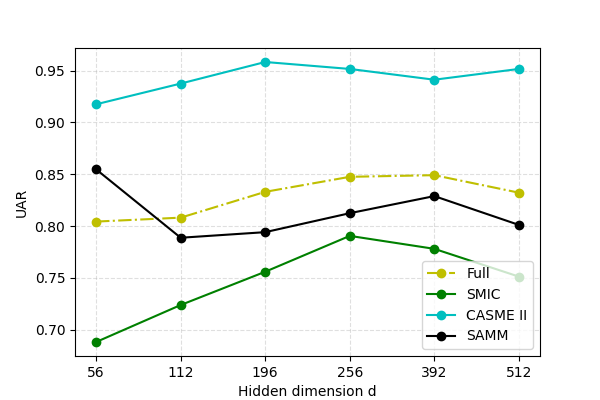}%
\label{dimension-fig_second_case}}
\caption{We research how the number of dimensions affects the accuracy of composite datasets-SMIC, SAMM and CASME II. The composite datasets' Unweighted F1-score (UF1) and Unweighted Average Recall (UAR) performance are reported.}
\label{dimension_number}
\end{figure}

\begin{figure}[!t]
\centering
\subfloat[]{\includegraphics[width=0.48\linewidth]{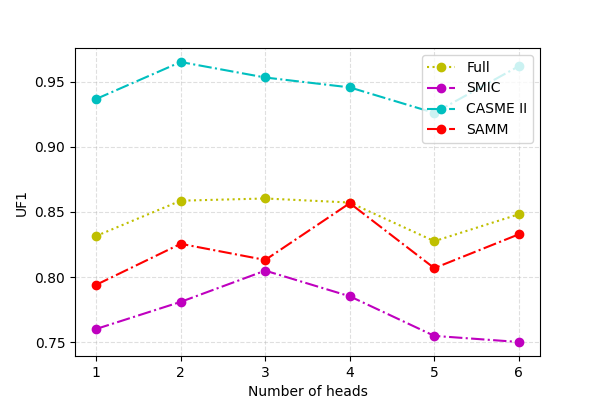}%
\label{heads-fig_first_case}}
\subfloat[]{\includegraphics[width=0.48\linewidth]{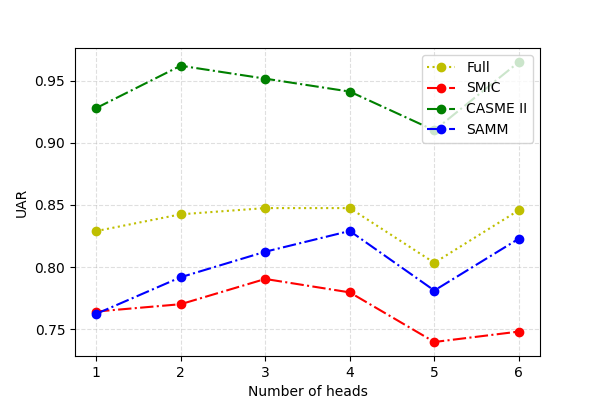}%
\label{heads-fig_second_case}}
\caption{We investigate the effects of the transformer layer's head count on accuracy in composite datasets-SMIC, SAMM and CASME II. The composite datasets' Unweighted F1-score (UF1) and Unweighted Average Recall (UAR) performance are reported.}
\label{heads_number}
\end{figure}
We conducted experiments to study the impact of different block sizes on the overall accuracy in composite micro-expression datasets, including SMIC, SAMM, and CASME II. The block size refers to the size of the facial regions considered in the HTNet model. We varied the bottom-level block size from 5 to 10, with the middle-level block size being twice the bottom-level size, and the top-level block size being four times the bottom-level size. The results, reported in Table \ref{impacts_of_block_size}, indicate that the choice of block size significantly affects the model's performance. Smaller block sizes may lead to subpar performance as they might miss some crucial facial parts, while larger block sizes generally perform better. However, if the block size becomes too large, some of the facial areas may overlap, such as the left-eye area overlapping with the right-eye area, which can negatively impact the model's performance. Therefore, finding an optimal balance between block sizes is crucial to ensure the best performance in micro-expression recognition tasks.

 \begin{table*}[t]
\centering
\small
\caption{ Study the impacts of the number of transformer layers on composite datasets. The composite datasets' Unweighted F1-score (UF1) and Unweighted Average Recall (UAR) performance are reported.}
\begin{tabular}{c| c c c c c c    }
\hline
\#Number of transformer layers & (2,2,2) & (2, 2, 4)& (2, 2, 6)& (2, 2, 8)& (2, 2, 10)& (2, 2, 12)\\
\hline
Full UF1   &0.8105 &0.8297  & 0.8316   & \textbf{0.8603} & 0.8464 & 0.8499\\
Full UAR  &0.7848 &0.8029  & 0.8183  & \textbf{0.8475}   & 0.8376 & 0.8207 \\
SMIC UF1   &0.748 &0.7247  & 0.744   & \textbf{0.8049} & 0.76 & 0.7561\\
SMIC UAR  &0.7363 &0.7065  & 0.734  & \textbf{0.7905}   & 0.756 & 0.7345 \\
SAMM UF1   &0.7984 &0.8137  & 0.7732   & 0.8131 & 0.8571 & \textbf{0.8682}\\
SAMMUAR  &0.75 &0.7849  & 0.7742  & 0.8124   & \textbf{0.8453} & 0.8419 \\
CASME II UF1   &0.8845 &0.9647  & \textbf{0.9722}   & 0.9532 & 0.9492 & 0.95\\
CASME II UAR  &0.8588 &0.9554  & \textbf{0.9658}  & 0.9516   & 0.9346 & 0.9412 \\
\hline
\#Params(millions)  &149.57  & 245.88  & 342.20 & 438.51 & 534.82 & 631.13  \\
\#Training time (Seconds)  &3767  & 4836  &  5906 & 6942 & 8085 & 9147\\
\hline
\end{tabular}
 \label{number-of-transformer-layers}
\end{table*}

\subsubsection{Impacts of dimensions}
We research how the number of dimensions affects the accuracy of composite datasets -SMIC, SAMM and CASME II. The Unweighted F1-score (UF1) and Unweighted Average Recall (UAR) performance of composite datasets are reported in Fig.\ref{dimension_number}. The smaller hidden dimension has worse performance because the small hidden dimension is hard to encode the optical flow feature map. However, using a larger hidden dimension will lead to overfitting problems. The full UF1 score for hidden dimension 256 has the best performance, around 0.86. 
\subsubsection{Impact of number of heads}
 We investigate the effects of the transformer layer's head count on accuracy in composite datasets- SMIC, SAMM and CASME II. The Unweighted F1-score (UF1) and Unweighted Average Recall (UAR) performance of composite datasets are reported in Fig.\ref{heads_number}. In Fig.\ref{heads_number}(a), it demonstrates that three heads in transformer layers will have the best performance. 

\subsubsection{Impact of different number of transformer blocks}
Table \ref{number-of-transformer-layers} presents an investigation into the impact of the number of transformer layers on the accuracy of our HTNet model, as evaluated using several datasets. To explore the effect of the number of transformer layers in the top-level, we vary the count from 2 to 12. As observed, an increase in the number of transformer layers leads to a corresponding growth in model parameters and training time. A smaller number of transformer layers may have limited capacity to effectively encode optical flow features. Conversely, a larger number of layers can introduce overfitting issues, impacting the model's performance.Our experiments indicate that employing two transformer layers in the bottom-level and middle-level, along with a moderate number of layers in the top-level, results in the best performance for our HTNet model.

\begin{figure*}[!t]
\centering
\subfloat[]{\includegraphics[width=0.33\linewidth]{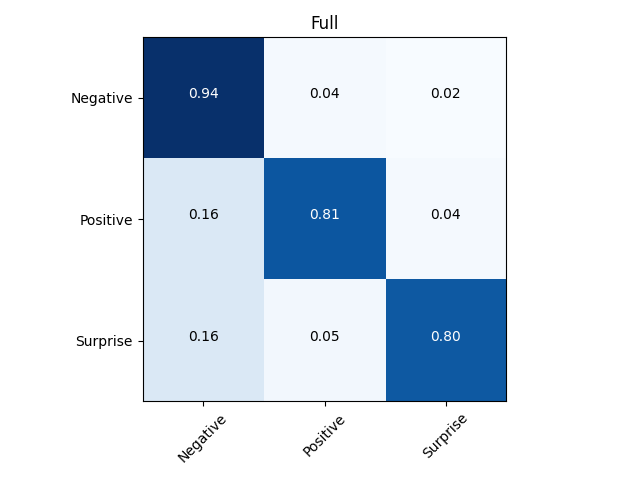}%
\label{fig_first_case}}
\subfloat[]{\includegraphics[width=0.33\linewidth]{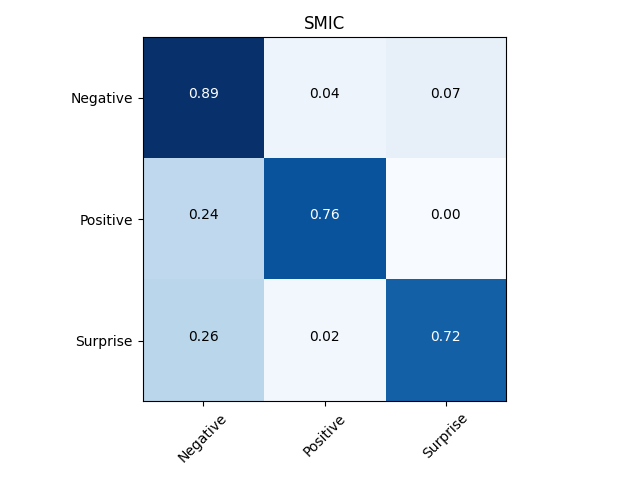}%
\label{fig_second_case}}
\subfloat[]{\includegraphics[width=0.33\linewidth]{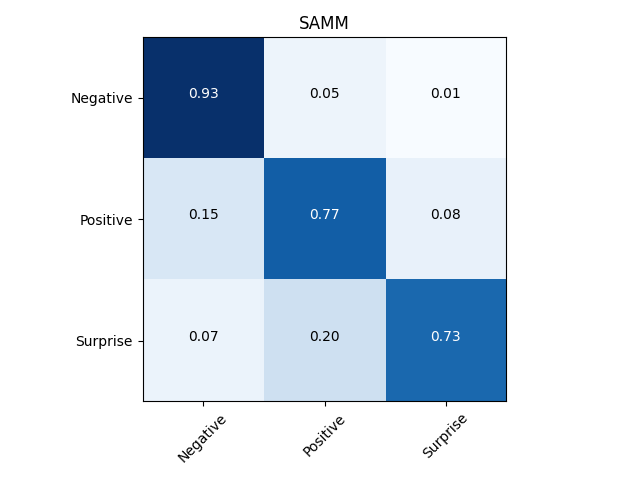}%
\label{fig_third_case}}
\hfill
\subfloat[]{\includegraphics[width=0.33\linewidth]{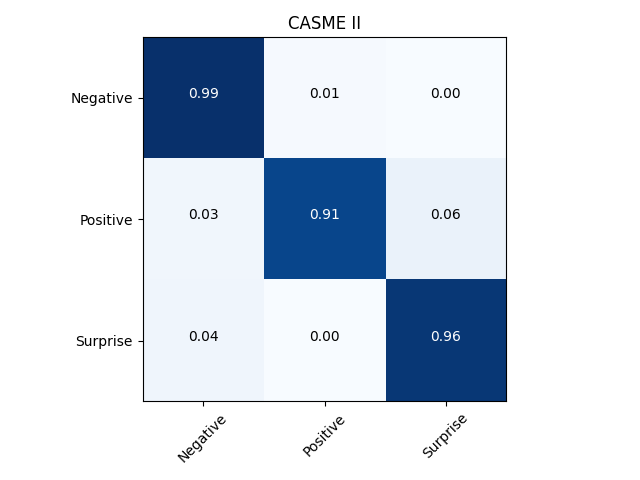}%
\label{fig_fourth_case}}
\subfloat[]{\includegraphics[width=0.33\linewidth]{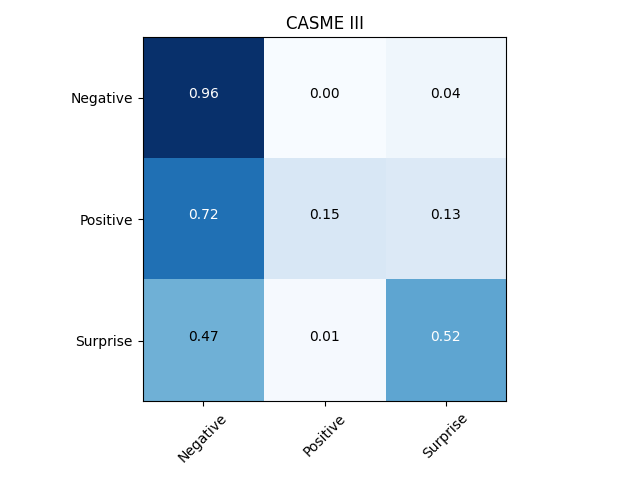}%
\label{fig_fifth_case}}
\caption{The confusion matrix for the proposed HTNet on the
composite database-SMIC, SAMM and CASME II and CASME III dataset using 3 classes.}
\label{fig_combine_confusion_matrix}
\end{figure*}
\section {Qualitative and quantitative analysis of our proposed method}

Figure \ref{fig_combine_confusion_matrix} (a) displays the confusion matrices of HTNet, presenting the accuracy achieved for each emotion category in the full composite datasets. Notably, HTNet attains accuracies of 0.94, 0.81, and 0.80 for the negative, positive, and surprise categories, respectively, in these datasets (SMIC, SAMM, and CASME II).The high accuracy for the negative category can be attributed to the larger number of training samples available for this class in the three datasets. However, challenges arise in distinguishing between the surprise and negative emotion categories in the SMIC and SAMM datasets. The limited number of training samples in these datasets may result in occasional misclassifications between these two categories.On the other hand, the CASME II dataset provides more precise apex frames, leading to more accurate optical flow feature maps that better capture facial muscle motions. Consequently, the negative, positive, and surprise categories in the CASME II dataset achieve relatively high accuracies, each exceeding 90\%. Misclassifications are minimal, further validating the efficacy of HTNet on this dataset.Additionally, the SMIC dataset employs lower frame rates for capturing images, which introduces background noise such as flickering lights, shadows, and variations in illumination. These factors can impact the accuracy of classifying positive and surprise emotions in the SMIC dataset, resulting in accuracies of 76\% and 72\%, respectively.

Figure \ref{fig_combine_confusion_matrix} (e) presents the confusion matrix obtained from this cross-validation. From the confusion matrix, we observe that the negative emotion category achieves the highest accuracy among the three emotion categories. This result is consistent with the performance observed in the composite datasets, where the negative category also exhibited the highest accuracy.On the other hand, the positive emotion category obtains the lowest accuracy, approximately 0.15. Most of the positive samples are misclassified as negative. In contrast, in the composite datasets, only a small portion of samples are misclassified as negative.

\section {Conclusion}
In this paper, we propose a hierarchical transformer architecture to extract important four facial areas features for micro expression recognition. Compared with previous deep learning methods, one of the key strengths of the HTNet model is its ability to model long-term temporal dependencies, which is crucial for accurate recognition of micro-expressions that often occur within a short timeframe. The hierarchical architecture of the model allows for the extraction of features at multiple temporal scales, which can improve recognition accuracy by capturing both short-term and long-term dynamics in the expression. In addition, HTNet model is based on the Transformer architecture, which has demonstrated state-of-the-art performance in natural language processing tasks. This architecture is well-suited for micro-expression recognition because it allows for the modeling of complex dependencies between different parts of the facial expression, while also being computationally efficient. 

Despite these strengths, HTNet model requires a large amount of training data, which may be difficult to obtain in some scenarios. The model may also be sensitive to variations in lighting, pose, and other environmental factors that can affect the appearance of micro-expressions. 

To address these limitations, future work can explore ways to improve the efficiency and robustness of the  HTNet model. Data augmentation techniques can be used to increase the amount of training data and improve the model's ability to generalize to different environmental conditions. Transfer learning can also be used to leverage pre-trained models on related tasks and improve the efficiency of the  HTNet model. With further development and optimization, the HTNet model may prove to be a valuable tool for accurately recognizing micro-expressions in various applications, including lie detection, emotion recognition and mental health assessment.

\section*{Acknowledgments}
I would like to thank my supervisors for valuable discussion, help and support. This work was partially supported by Australian Government Research Training Program Scholarship.





\bibliographystyle{elsarticle-num}
\bibliography{books}






\newpage
\setlength\intextsep{0pt} 
    \begin{wrapfigure}{l}{0.13\textwidth}
        \centering
        \includegraphics[width=0.15\textwidth]{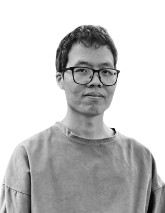}
\end{wrapfigure}
\noindent \textbf{Zhifeng Wang} graduated with an MS degree in computer vision and machine learning from the Australian National University in 2021. He is presently pursuing a Ph.D. at the Australian National University's College of Engineering and Computer Science in Canberra, ACT, Australia. His study focuses on computer vision and machine learning, particularly deep learning for facial and emotion identification.

\subsection*{  } 
\setlength\intextsep{0pt} 
    \begin{wrapfigure}{l}{0.13\textwidth}
        \centering
        \includegraphics[width=0.15\textwidth]{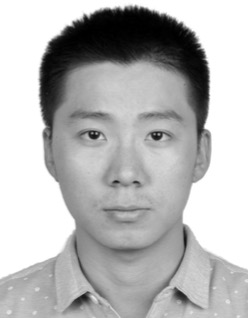}
\end{wrapfigure}
\noindent \textbf{Kaihao Zhang} (Graduate Student Member, IEEE) is currently pursuing the Ph.D. degree with the College of Engineering and Computer Science, The Australian National University, Canberra, ACT, Australia. He has more than 20 referred publications in international conferences and journals, including CVPR, ICCV, ECCV, NeurIPS, AAAI, ACMMM, IJCV, IEEE TRANSACTIONS ON IMAGE PROCESSING (TIP), and IEEE TRANSACTIONS ON MULTIMEDIA (TMM). His research interests focuses on
computer vision and deep learning
\subsection*{  } 
    \setlength\intextsep{0pt} 
    \begin{wrapfigure}{l}{0.13\textwidth}
        \centering
        \includegraphics[width=0.15\textwidth]{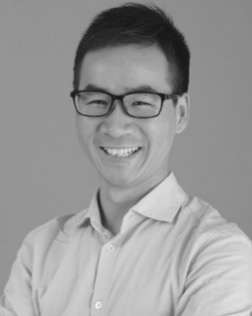}
    \end{wrapfigure}
    \noindent \textbf{Wenhan Luo} is an associate professor and PhD advisor in Sun Yat-sen University, conducting research on trustworthy AI and creative AI. Before being a faculty member in university, he worked as an applied research scientist at Tencent, solving real-world problems using computer vision and machine learning techniques. Prior to Tencent, he worked for Amazon (A9) in Palo Alto, California, where he developed deep models for better visual search experience. Before that, he worked as a research scientist in Tencent AI Lab.  He received the Ph.D. degree from Imperial College London, UK, 2016, M.E. degree from Institute of Automation, Chinese Academy of Sciences, China, 2012 and B.E. degree from Huazhong University of Science and Technology, China, 2009. He have published over 60 peer-reviewed papers, including over 40 of them published in top-tier conferences and journals, like ICML, CVPR, ICCV, ECCV, AAAI, ACL, ACMMM, ICLR, TPAMI, AI, IJCV, TIP, and 2 of them are ESI highly cited papers. He received the CVPR 2019 Best Paper Nominee and was awarded the 2022 ACM China Rising Star Award (Guangzhou Chapter).

\subsection*{  } 
    \setlength\intextsep{0pt} 
    \begin{wrapfigure}{l}{0.13\textwidth}
        \centering
        \includegraphics[width=0.15\textwidth]{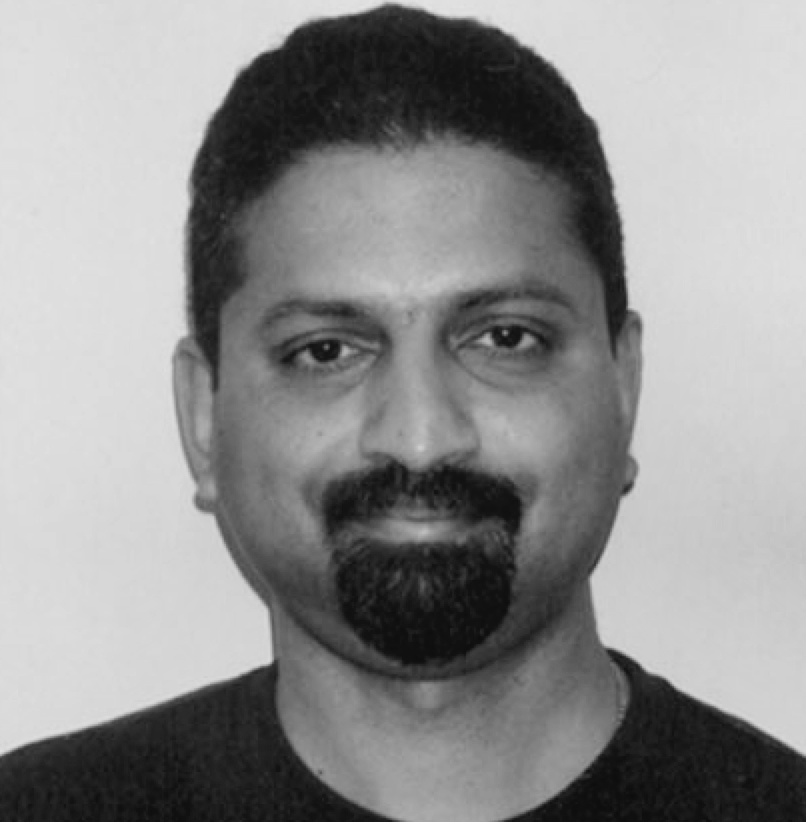}
    \end{wrapfigure}
    \noindent \textbf{Ramesh Sankaranarayana}  received the ME degree from the Indian Institute of Science, India, and the PhD degree from the University of Alberta, Canada. He is currently the associate director (Educational partnerships), Research School of Computer Science, Australian National University, Canberra, Australia. He research interests include information retrieval, human-centered computing and software engineering. He is a member of the ACM.
\end{document}